\documentclass[10pt, a4paper]{article}
\pdfoutput=1
\usepackage[utf8]{inputenc}
\usepackage[pdftex]{graphicx}
\usepackage{lrec2006}
\usepackage{times}
\usepackage{url}
\usepackage{latexsym}

\usepackage[english]{babel}

\usepackage{placeins}

\usepackage{amsmath}
\usepackage{amsfonts}
\usepackage{amssymb}

\usepackage{amsthm}

\usepackage{moreverb}
\usepackage{listings}
\usepackage{caption}
\usepackage{subcaption}
\usepackage[pdftex]{graphicx}

\graphicspath{{dia/}}

\usepackage{acronym}
\acrodef{SL}{Sign Language}
\acrodef{SLR}{Sign Language Recognition}
\acrodef{SLP}{Sign Language Processing}
\acrodef{MG}{Manual Gesture}
\acrodef{NMG}{Non-Manual Gesture}
\acrodef{LSF}{French Sign Language}
\acrodef{NLP}{Natural Language Processing}
\acrodef{ST}[SplT]{Simple Tree}
\acrodef{SG}{Solution Graph}
\acrodef{CSP}{Constraint Satisfaction Problem}
\acrodef{CFG}{Context-Free Grammar}
\acrodef{PSG}{Phrase Structure Grammar}

\usepackage[usenames,dvipsnames]{color}

\title{A hybrid formalism to parse Sign Languages}

\name{Rémi Dubot, Christophe Collet}

\address{ IRIT \\
Université de Toulouse \\
France \\
dubot@irit.fr, collet@irit.fr\\}

\abstract{
Sign Language (SL) linguistics is dependent on the expensive task of annotation. 
Some automation is already available for low-level information (eg. body part tracking) and the lexical level has shown significant progresses. 
The syntactic level lacks annotated corpora as well as complete and consistent models. 
This article presents a solution for the automatic annotation of SL syntactic elements. It exposes a formalism able to represent both constituency-based and dependency-based models. The first enables the representation of structures one may want to annotate, the second aims at fulfilling the holes of the first. 
A parser is presented and used to conduct two experiments to test the solution. One experiment is on a real corpus, the other is on a synthetic corpus. 
}

\begin{document}
\maketitleabstract

\acresetall
\section{Introduction}

To study \acp{SL}, linguists need annotations. 
Currently, corpus annotation is done manually, it is time-consuming and suffers difficulties with inter and intra-annotator reliability. 
For this reason, efforts are carried out to automatize the annotation process. 
Early efforts focused on the very low-level non-linguistic information: body part tracking, activity detection. They finally reached the base of the linguistic level: detection of sign phases~\cite{workshop_signs_2011}, sub-lexical~\cite{cooper_sign_2012} and lexical units~\cite{curiel_sign_2013}. Work on this last level has focused on manual gestures. The only exceptions were attempts to remove ambiguity on some lexical signs with the help of \acp{NMG}~\cite{paulraj_extraction_2008} or detection of \ac{NMG}~\cite{yang_combination_2011,neidle_method_2009}. 
Now is the time to address the annotation of supra-lexical features. But when it comes to syntactic features, it is not possible to ignore the \acp{NMG} anymore.

The syntax \acp{SL} is complex and different from vocal languages\cite{cuxac_langue_2000,dubuisson_grammaire_1999,bouchard_grammar_1995,bouchard_sign_1996}. They use the multiplicity and the spatial abilities of the available articulators. 
It results non-sequential productions with complex temporal, spacial and articulatory synchronizations. 
The syntactic models developed for the processing of vocal languages are deeply based on the sequentiality of lexical units. Consequently, the processing of \ac{SL} syntax requires the invention of new models or, at least, to deeply rethink and adapt the existing ones. 

A recognition system always has an internal representation of the phenomena to recognize. 
However, there are multiple manners to obtain such a representation. From one extreme to another, it can be expert knowledge formalized into a model or it can be results of uninformed automatic learning on real data. The first requires experts to formalize a complete and consistent model from their knowledge. The second requires massive data and computer calculation. For the syntax of \acp{SL}, neither is available. The expert knowledge is sparse and sometimes inconsistent. 
Annotated \ac{SL} corpora are too small and too heterogeneous for uninformed learning. 

Our goal is to develop tools for the semi-automatic annotation. 
The general approach we adopt is to use supra-lexical/syntactic models for the annotation. It targets two objectives. 
First, it aims at producing annotations for all the structures of the model. 
Second, it aims to enhance the lower levels. 
Indeed, such models can improve two aspects of the quality of the lexical recognition: the results, by re-scoring the lexical candidates, and the efficiency, by informing the lexical layer and thereby reducing the search space. 
The models are used to propagate the information of the low-level detections.

This article exposes elements in favor of a hybrid parsing of \acp{SL}. 
It presents a formalism able to represent constituency-based structures as well as dependency-based structures. This formalism has been created to represent models combining transfered linguistic knowledge and automatically learned dependencies. The feasibility is demonstrated with a parser in two experiments. 
First, the parser is run on excerpts of the Dicta-Sign Corpus with a model composed of five structures. 
Second, synthetic dependency grammars are used to parse synthetic corpora. 

Such a hybrid formalism is the solution we found for the lack of annotated corpora and the incompleteness of the available models.  
We aim at enabling the use of incomplete models transfered from the linguistic knowledge with learned data. 

This work tries to avoid hypotheses that would simplify \ac{SL} processing by making \acp{SL} closer of vocal languages but would be unrealistic. 
In particular, it makes no assumptions such as the predominance of the hands over the other articulators or the existence of a sequential skeleton of the \ac{SL} locutions. It is based on the ideas introduced by Filhol~\cite{filhol_descriptive_2009} to represent structures with the minimal constraints that make them recognizable. This approach enable to naturally represent the complex temporal synchronization mechanisms~\cite{filhol_combining_2012} of \ac{SL} simultaneity~\cite{vermeerbergen_simultaneity_2007}. 

This document is structured as follow. 
It starts with the presentation of the example used all along the article. 
The formalism is described jointly with its usage for constituency-based structures. 
The representation of dependency structures comes next. 
After the formalism, the parsing is presented with its general characteristics but without details on its internal algorithm. 
The last part presents the two experiments, their results and an analysis. 

\begin{figure*}[h!tb]
\centering
\begin{subfigure}[c]{0.125 \textwidth}
	\centering
	\includegraphics[scale=0.25]{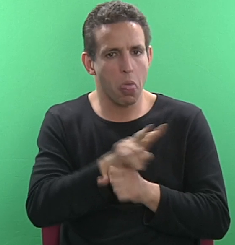}  
	\caption{VISIT}
\end{subfigure}
\begin{subfigure}[c]{0.125 \textwidth}
	\centering
	\includegraphics[scale=0.25]{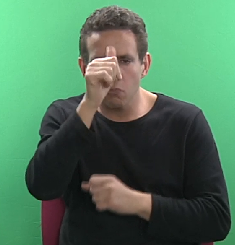}  
	\caption{MUSEUM}
\end{subfigure}
\begin{subfigure}[c]{0.125 \textwidth}
	\centering
	\includegraphics[scale=0.25]{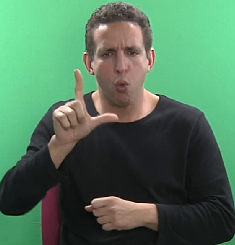}  
	\caption{L}
\end{subfigure}
\begin{subfigure}[c]{0.125 \textwidth}
	\centering
	\includegraphics[scale=0.25]{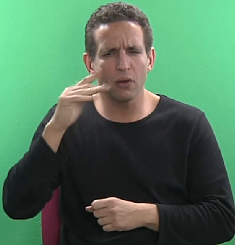}  
	\caption{FORGET}
\end{subfigure}
\begin{subfigure}[c]{0.125 \textwidth}
	\centering
	\includegraphics[scale=0.25]{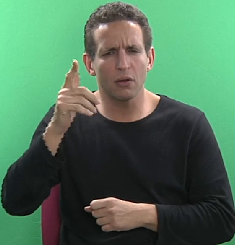}  
	\caption{pointer}
\end{subfigure}
\begin{subfigure}[c]{0.125 \textwidth}
	\centering
	\includegraphics[scale=0.25]{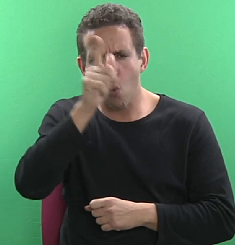}  
	\caption{LOUVRE}
\end{subfigure}
\begin{subfigure}[c]{0.125 \textwidth}
	\centering
	\includegraphics[scale=0.25]{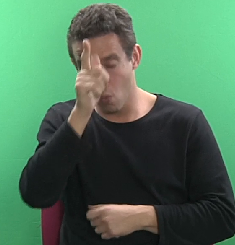}  
	\caption{LOUVRE}
\end{subfigure}
\caption{Decomposition of the excerpt}
\label{chrono}
\end{figure*}

\section{Formalism description}

The first step toward the automatic annotation is the formal representation of a model. 
The representation we propose is similar to \acp{CFG} in that it is a derivational grammar. But it differs from \acp{CFG} on three fundamental points. First, the right-hand side of a production rule is not a string of units but a set of units. Second, it introduces the possibility to express constraints between all the units of a production rule. Third, in \acp{CFG}, the left-hand side of a production rule is non-terminal symbol. We have no such thing as non-terminal and terminal symbols. We have instead detectable and non-detectable units, and both can be atomic (terminal) or not. 

We target the representation of two types of models. 
In the first, a production rule represents a relation of constituency. It comes from the \acp{PSG} of Chomsky~\cite{chomsky_syntactic_1957}. In the second, a production rule represents a relation of dependency. It comes from the dependency grammars of Tesnière.

\subsection{Constituency structures}
\subsubsection{Example presentation}
We illustrate the description of the formalism with the construction of a constituency-based model from an excerpt of a real corpus. 

The excerpt comes from the \ac{LSF} part of the Dicta-Sign corpus~\cite{efthimiou_dicta-sign:_2010} which is composed of spontaneous dialogs performed by deaf signers. In this excerpt, the informant relates a memory of a journey in Paris visiting the Louvre museum with a friend. 
In the studied part, he explains to his interlocutor the purpose of the journey --to visit the Louvre-- and checks that they share the same sign for Louvre. Figure~\ref{chrono} summarizes the excerpt with a sequence of pictures.

\subsubsection{Pattern decomposition}

\begin{figure*}[h!tb]
	\centering
	\begin{subfigure}[c]{0.27\textwidth}
		\centering
		\includegraphics[scale=0.5]{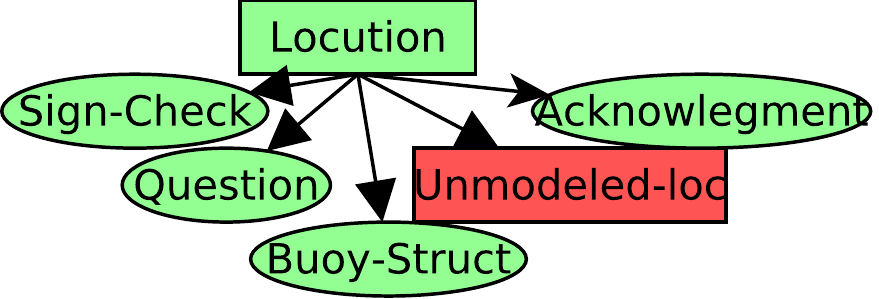}
		\caption{Locution}
		\label{fig:locution}
	\end{subfigure}
	\begin{subfigure}[c]{0.20\textwidth}
		\centering
		\includegraphics[scale=0.5]{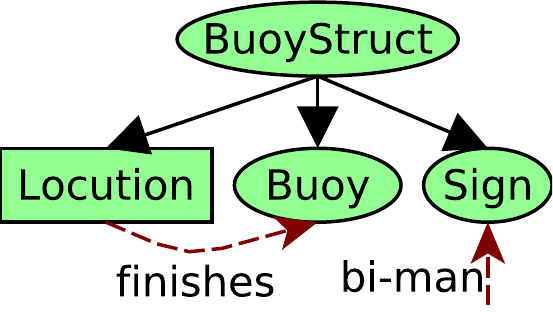}  
		\caption{Buoy}
		\label{fig:buoy-struct}
	\end{subfigure}
	\begin{subfigure}[c]{0.142\textwidth}
		\centering
		\includegraphics[scale=0.5]{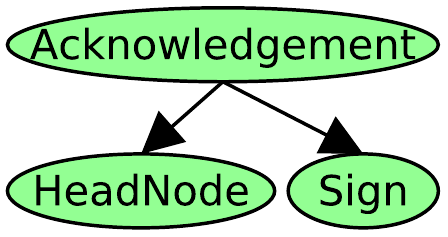}  
		\caption{Ack}
		\label{fig:ack-struct}
	\end{subfigure}
	\begin{subfigure}[c]{0.135\textwidth}
		\centering
		\includegraphics[scale=0.5]{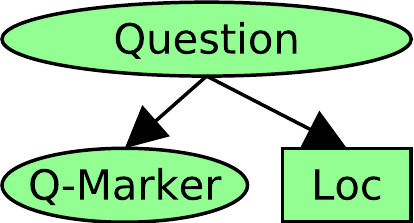}  
		\caption{Question}
		\label{fig:question-struct}
	\end{subfigure}
	\begin{subfigure}[c]{0.23\textwidth}
		\centering
		\includegraphics[scale=0.5]{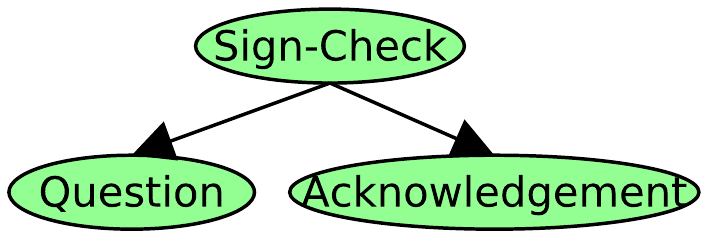}  
		\caption{"Sign Check"}
		\label{fig:check-struct}
	\end{subfigure}
\caption{Example of model with 4 patterns (b, c, d \& e) }
\label{model}
\end{figure*}

We call pattern a rule representing how a unit comes with others. It is similar to the production rules of \acp{CFG}. We usually draw these patterns as trees as shown in figure~\ref{model}. 
In the present formalism, we make each pattern correspond to a unit (the inverse is false, it is not an equivalence relation). 
Consequently, a unit can be the root of at most one pattern for a given model. 
An atomic unit can be associated to a pattern with only a root. 
It is the single assumption make about units and patterns in a model. 
Aside from this, everything is possible. Units can appear several times in the same pattern. Patterns can be recursive, mutually recursive, etc.   

The model we are about to introduce contains four patterns observed in the excerpt: a buoy pattern, a ``sign check" pattern, a question pattern, and an acknowledgment pattern. 
These patterns are examples and do not rely on a strong linguistic basis. Stronger models remain to be developed with linguists. 

The patterns are described in terms of constituents as shown in figure~\ref{model}. Their internal arrangement is then described with constraints (section~\ref{cons}). 

The first described pattern is a buoy~\cite{liddell_grammar_2003}. 
It is visible in figure~\ref{chrono}, the left hand of the bi-manual sign TO-VISIT (fig.~\ref{chrono}(a)) is maintained all along the excerpt. 
The pattern is decomposed into three sub-elements: two signs and one locution. 
The second pattern is an acknowledgment. It happens in figure~\ref{chrono} (g). It is decomposed into two sub-elements: a head node and a sign. 
The third pattern is a question. It also happens in figure~\ref{chrono} (g), but is less clear on this snapshot. It is decomposed as a marker (eyebrows up) and a locution. 
The ``sign check" is a question and an acknowledgment.

As shown in figure~\ref{model}, the pattern decomposition can be easily represented as a tree. 
The sub-elements are patterns which can be decomposed themselves or can be considered atomic in the model. Edges represent a relation of constituency. 
In a decomposition, multiple elements can be instances of a same pattern. 
When defining a model, one may need to introduce the same pattern multiple times in a same decomposition.  
This fact is of particular importance as it highlights that an element, in a decomposition, does not represent a pattern but an instance. As a consequence, the name of a pattern is not sufficient to designate elements without ambiguity. It is therefore necessary to associate each instance with a role name. 

\subsubsection{Alternatives}
Patterns do not allow generalization as all their internal elements are mandatory. 
As patterns describe compositions, we define an other type of rule to explicitly express alternatives. 
The same restriction as for patterns applies to the use of a unit as root for an alternative. 
In the example model, we define a node \textit{Locution} as an alternative between the four patterns (figure~\ref{fig:locution}). Alternatives appear as rectangle nodes in figures~\ref{model} and \ref{fig:sg}.

\subsubsection{Constraints}
\label{cons}
Patterns and alternatives represent invariants in the composition. 
Invariants in the internal organization of the patterns are expressed with constraints. 

To come back to the example, we can extract several kinds of invariants. One may hypothesize that the sign beginning a buoy structure must be \textit{bi-manual} (figure~\ref{fig:buoy-struct}). 
Another may want to describe the temporal structure of the patterns (Buoy~\textit{finishes}~BuoyStruct, in figure~\ref{fig:buoy-struct}). 
It could also be useful to express global constraints, for instance constraints between one unit and all its descendants. 
All these invariants should be expressible formally. 

We represent temporal, spatial and articulatory invariants as constraints. 
The constraints restrain the possible values for the attributes of pattern instances. The attributes, their encoding, and the logic formalisms -- used to express the constraints -- are a whole. Their choice strongly impacts the model. 
This is the reason why the formalism has to be independent of the logics and attributes. 

Representing a complete model requires multiple logics, 
each addressing a different aspect: 
temporal, spatial, articulatory, etc. 
We showed examples of the temporal (\textit{finishes}) and articulatory (\textit{bi-manual}) aspects. 
In this article, we focus on the formalism to describe the model. For this demonstration, only temporal constraints are used. 

\subsection{Edges of the models}
\label{edges}
Developing a complete model is, at best, very hard. 
We consider two solutions to work with incomplete models. 
As this work is developed for semi-automatic annotation, the first solution is to transfer the charge to the human operator. 
Such a system would ask something like ``There might be a `Question' there, is there an `unmodeled-loc'? and which are its characteristics (attributes)?". 
This solution requires from the operator precisely what makes annotation difficult for humans: he/she is supposed to fulfill many attributes that are hard to measure for a human being. 
This problem leads to the second solution: coarse-grained models. 
Such models are not meant for the analysis of their results, they intend to produce a block with attribute values similar to what could have produced a complete model. 
Our solution combines these two approaches. 

When a model is incomplete, edge nodes appear which are used but not modeled.  
Such an edge is present in the example model as ``unmodeled-loc". 
The ``unmodeled-loc" represents locutions built using non-modeled structures. We have built an experimental coarse model based on the sequence of lexical signs (because the annotation was already existing). 
The results, as expected, are not good. Depending on how constrained we make the model, we have far too much false-negatives or false-positives. The sequence model does not work well with the overlapping units: it includes units we don't want included and vice-versa. 
We expect dependency-based models to constitute better coarse-grained models.  

\subsection{Dependency structures}
\label{deps}
For the dependency grammar part, we present the formalism with a model which makes several simplistic hypotheses. 
The example model divides the units in two types: \acp{MG} and \acp{NMG}, each one with its proper behavior. The units can represent a variety of forms: standard signs, other \acp{MG} (e.g. pointing \acp{MG}), facial gestures (e.g. qualifiers, quantifiers, modality markers), gaze gestures (e.g. references), etc. 
In \acp{SL}, articulatory constraints impact the syntactic level. 
Some units interact and some others are incompatible. 
In this example, the model emulates simplified articulatory interactions between its units:  
\begin{itemize}
\item \acp{MG} never overlap. This is a simplification as it excludes the representation of yet described phenomenons (e.g. buoy structures, Cuxac's situational-transfers~\cite{cuxac_langue_2000}). 
\item All \acp{NMG} can overlap. This is a simplification as some \acp{NMG} are articulatory impossible to produce simultaneously. 
\end{itemize} 

These simplifications allowed us to work with a slightly extended version of the Hays' formalism. 
Hays defines rules of the form $X(Y_{-n},...,Y_{-1},*,Y_1,...,Y_m)$ where $X$ and $Y_k$ are categories of units. Such a rule expresses that a unit of category $X$ takes the place of the star in a sequence of dependents of categories $Y_{-n}$ to $Y_m$. 
This formalism is sufficient to represent \acp{MG} (assuming the sequence simplification). 
But the \acp{NMG} requires to extend it, which is done with rules of the form $X(Y)$.

\begin{figure}
\centering
\includegraphics[scale=0.6]{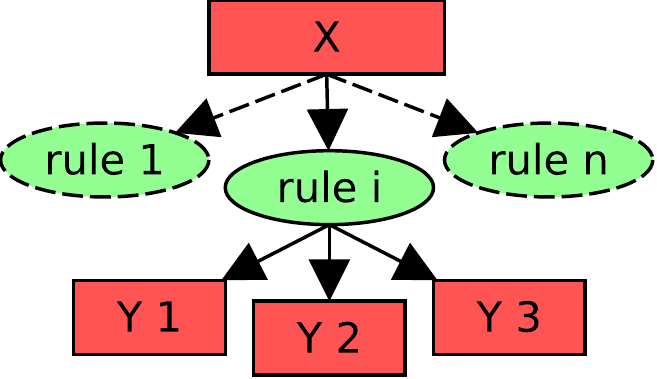}
\caption{Representation of a dependency}
\label{fig:deps}
\end{figure}

We have represented such dependency structures with the formalism with the construction shown in figure~\ref{fig:deps}. 
The categories are described as alternatives between rules. The rules are described as patterns. 
The constraints work exactly as for constituency-based structures.

\section{Parsing}

The purpose of this work is the semi-automatic annotation of structures of models. 
The first step toward this objective was to formalize the model to recognize. 
The next step is the recognition itself. 
We give here an outline only of the developed system. The detailed description will be the subject of a dedicated article. 

\begin{figure*}[htb]
	\centering
	\includegraphics[scale=0.6]{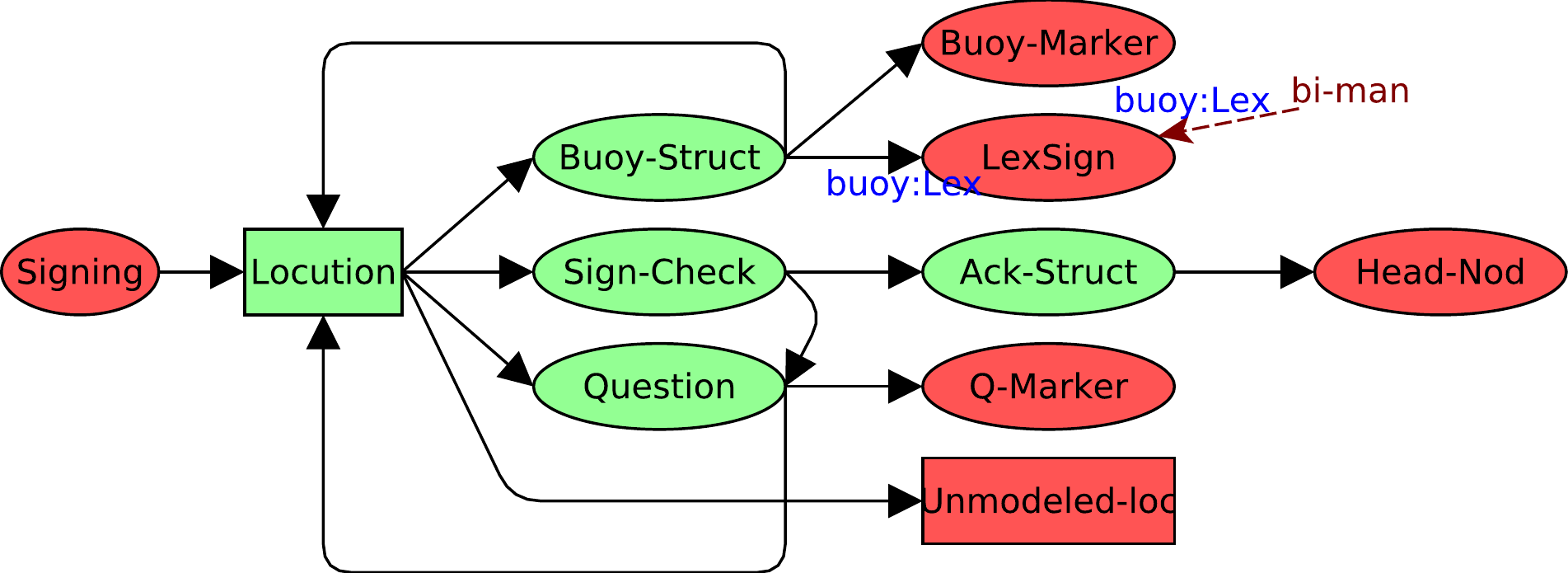}  
	\caption{Schematic view of the implicit graph associated to the example model}
	\label{fig:sg}
\end{figure*}

In addition to the formalized model, the parser needs an input to parse. 
This input is an annotation of a subset of the units of the model. 
Units of this subset (they can be either pattern or alternatives) are said to be detectable. 
Their annotation can originate from manual annotation or third party detectors. 
These detectable units appear in red in figures~\ref{fig:sg} and~\ref{fig:deps}. 
The parser is able to command the external detectors as it runs. In this mode, it does not receive the input annotation a priori, but works interactively with the detectors. This allows to inform the detectors of the context and therefore to reduce their search spaces. On the example, the parser asks to the ``Buoy-Marker" detector ``is there something between 201 and 212?". This allows to reduce the time interval the detector will process. 

The internal representation of the model in the parser is an AND/OR graph. This representation is called the implicit graph. 
Our work extends the ideas of Mahanti~\cite{mahanti_framework_2003} for the parsing. 
A unit identifying a pattern gives an \textit{AND} node and one identifying an alternative gives an \textit{OR} node. 
In the implicit graph, nodes represent patterns or alternatives but not instances. 
Figure~\ref{fig:sg} gives an example of an implicit graph for the example model. 
The implicit graph is used to generate an explicit graph. In this last graph, nodes represent instances. 

\begin{figure*}[htb]
	\centering
	\includegraphics[scale=0.5]{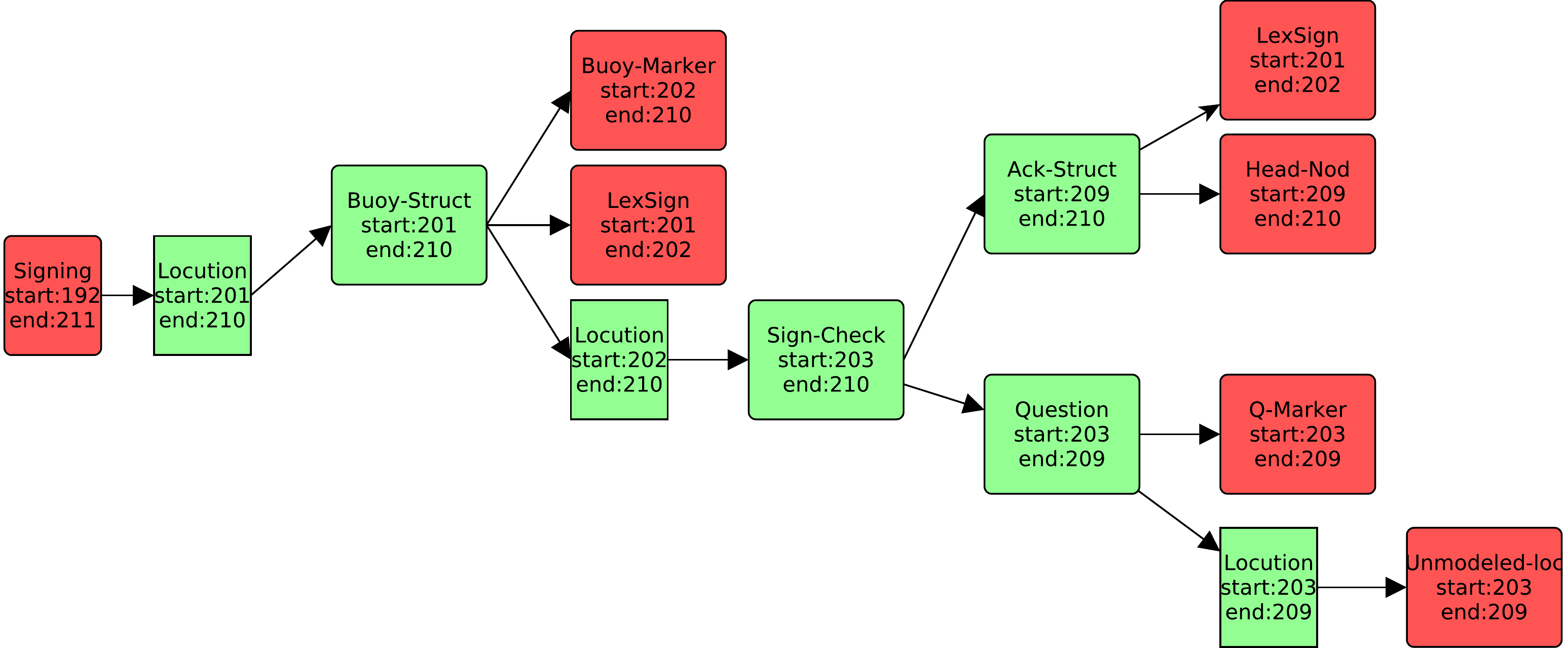}
	\caption{Example of solution graph}
	\label{fig:solution}
\end{figure*}

The parsing operation results in a set of graphs. 
Each graph is a solution. 
The figure~\ref{fig:solution} shows an example of graph output by the parser. 
The nodes represent occurrences either externally detected or internally inferred. 
The arcs correspond to constituency or dependency relations of the model. 

In a solution graph, each node has attributes. 
As the model can be under-constrained, there may be more than one solution. In particular, the resolution can find more than one acceptable value for attributes. 

The parser is currently top-down. It builds the solution graphs starting from a set of given roots. 
This set can be, for example, a set of pre-detected lexical unit occurrences resulting of a first pass of lexical recognition. It is how the parser process dependency-based models. It then builds trees top-down from each root and merges the trees when possible. It is therefore obvious than solution graphs can have multiple connected components. This occurs, for example, when a signer is interrupted by a question, answers quickly and then continues his/her speech. 
In the case of constituency-based models, the top-down parsing requires to introduce a detectable root. It is the function of the "Signing" unit in figure~\ref{fig:sg} which is detected with an activity detector.

In the models we developed, the set of attributes contains \emph{time-start} and \emph{time-end}. Their values make it easy to transform a solution graph into an annotation.

\section{Results}

The parser has been evaluated for constituency-based and dependency-based structures: the first on real annotations, the second on synthetic data. 
The results of the parser can be directly observed, quantitatively and qualitatively. 
The evaluation of the formalism itself is harder to produce. We propose an interpretation of the parser's results to understand what they say about the formalism. 

The parser has been run on several occurrences of the constituency-based structures. 
The external detectors were simulated with a manual annotation of the detectable units. 
But the small number of occurrences does not allow a quantitative evaluation. 
In particular, the evaluation corpus contains only one occurrence of a combination of the structures. 

We still produce a qualitative analysis of the results. 
The parser outputs numerous solutions: many false-positives and partial solutions. 
A simple ranking by the size of the solutions is efficient against the partial solutions. 

The false-positives can be classified in two categories: wrong hierarchical order and bad modeling of the lower levels of the syntax (discussed above, in section~\ref{edges}). 
The first could be addressed with recursive constraints on the compositions. For example constraints like ``the locution constituting a question cannot contain a question". Such a feature could be interesting for experiments on models. But in a context of semi-automatic annotation, we rather think that this type of false-positives must be resolved by a human expert. A system requiring this type of intervention of the operator is still of good help: it reduces the work in the task of selecting the right hierarchical organization. This uses the expertise of the operator for high-level problems. 
The second type of false-positives comes from the difficulty we met in modeling the syntactic structures of low-level. It is the reason why we developed the dependency part of our formalism. 

To evaluate the parser on dependency grammars, we have built a synthetic corpus. The idea behind this is to test the parser against bigger inputs. 
To generate this corpus, we used the model presented in the section~\ref{deps} 

Our generator starts with the random generation of dependency grammars. It then generates random phrases following the grammars. In the absence of measures on annotations, the models were parametrized arbitrarily. 
The corpus has 5000 grammars with 1 phrase each. All grammars have 20 categories. Every category has 3 to 4 rules each. Rules for non-manual categories have exactly one dependent. For manual categories, sizes have a uniform distribution on $\left[0, 4\right]$.  

\begin{figure}
\centering
\includegraphics[width=1\columnwidth]{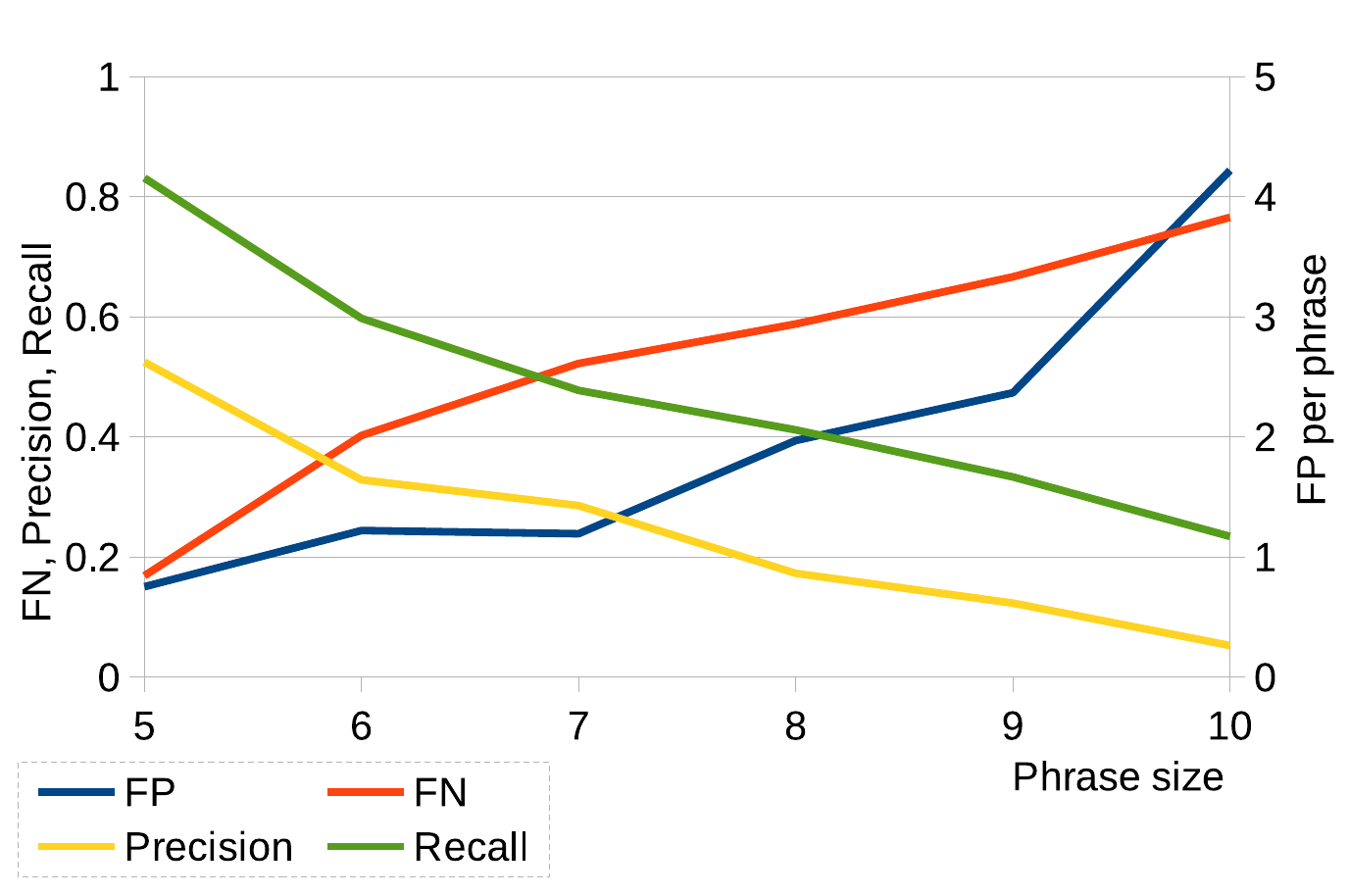}
\caption[]{Evaluation on dependency grammars}
\label{fig:synth-res}
\end{figure}

The results of the parsing on the synthetic corpus are visible in figure~\ref{fig:synth-res}. 
The results are classed by phrase size. 
We have an average of 1 to 4 false-positives per phrase. It gives a precision of 52\% to 5\%. It is hard to draw conclusion from this result as it depends on the parameters chosen at the grammar generation. 
The recall of 83\% to 23\% is much more interesting. It validates the computability of the parsing. 

\section{Conclusion}

The formalism of this article showed its ability to represent structures based on constituency as well as dependency relations. 
It has been done without assumptions on the sequentiality of lexical units nor on the predominance of the manual gestures. 
Instead, it uses constraints to describe invariants on the composition of the structures and on their temporal organization. 
We showed that these descriptions allow the detection of the structures. 
The dependency parsing shows promising results as a coarse model. 
This should ease the use of constituency-based structures by disassociating them from the complete model requirement. 
However, the articulation between the two paradigms in one model remains to be developed. 
For now, the solution is to have two separated models, one per paradigm. 
The dependency-based model is used when a non-modeled pattern is reached. 
At this time, the human operator decides if the pattern is present and what solution of the dependency parsing will act as the occurrence of the non-modeled pattern.

This work, in its current state, is restricted by some limitations of the generative grammars. 
But it already avoids the problem of designing a model with a unique root for dependency grammars. This is critical in our context of semi-automatic annotation, as our goal is to enable the detection of structure occurrences, not to produce an interpretable syntactic tree. 
Unfortunately, the parser is still top-down, and consequently, the constituency-based grammars still need a root. There are plans to modify the current parser to drop the top-down mechanism. 
This will enable the parser to accept non-rooted models.


To go further in the direction of automatic annotation, several points need to be worked on. First, one will have to build (manually or automatically) a dependency grammar compliant with a real \ac{SL}. The formalism and the parser can manage models of dependency grammars much more complex than one presented above. 

The formalism and the parser do not represent uncertainty. But there are good candidates to introduce uncertainty representation in the existing parser such as fuzzy-CSPs. This extension will certainly improve greatly the results but will also have a computational cost. 

\newpage

\pagebreak

\bibliographystyle{lrec2006}
\bibliography{MyLibrary}

\begin{thebibliography}{}

\bibitem[\protect\citename{Bouchard and Dubuisson}1995]{bouchard_grammar_1995}
Denis Bouchard and Colette Dubuisson.
\newblock 1995.
\newblock Grammar, order \& position of wh-signs in quebec sign language.
\newblock {\em Sign Language Studies}, 87(1):99–139.

\bibitem[\protect\citename{Bouchard}1996]{bouchard_sign_1996}
Denis Bouchard.
\newblock 1996.
\newblock Sign languages \& language universals: The status of order \&
  position in grammar.
\newblock {\em Sign Language Studies}, 91(1):101–160.

\bibitem[\protect\citename{Chomsky}1957]{chomsky_syntactic_1957}
Noam Chomsky.
\newblock 1957.
\newblock {\em Syntactic structures}.
\newblock {Mouton\&Co}, La Haye.

\bibitem[\protect\citename{Cooper \bgroup et al.\egroup
  }2012]{cooper_sign_2012}
Helen Cooper, Eng-Jon Ong, Nicolas Pugeault, and Richard Bowden.
\newblock 2012.
\newblock Sign language recognition using sub-units.
\newblock {\em Journal of Machine Learning Research}, 13:2205--2231.

\bibitem[\protect\citename{Curiel and Collet}2013]{curiel_sign_2013}
Arturo Curiel and Christophe Collet.
\newblock 2013.
\newblock Sign language lexical recognition with propositional dynamic logic.
\newblock In {\em Proceedings of the 51st Annual Meeting of the Association for
  Computational Linguistics}, volume~2, page 328–333.

\bibitem[\protect\citename{Cuxac}2000]{cuxac_langue_2000}
Christian Cuxac.
\newblock 2000.
\newblock {\em La langue des signes française ({LSF):} les voies de
  l'iconocité}.
\newblock Ophrys.

\bibitem[\protect\citename{Dubuisson \bgroup et al.\egroup
  }1999]{dubuisson_grammaire_1999}
Colette Dubuisson, Lynda Lelièvre, and Christopher Miller.
\newblock 1999.
\newblock {\em Grammaire descriptive de la {LSQ.}}
\newblock Université du Québec à Montréal.

\bibitem[\protect\citename{Efthimiou \bgroup et al.\egroup
  }2010]{efthimiou_dicta-sign:_2010}
Eleni Efthimiou, Stavroula-Evita Fotinea, Thomas Hanke, John Glauert, Richard
  Bowden, Annelies Braffort, Christophe Collet, Petros Maragos, and François
  Goudenove.
\newblock 2010.
\newblock {DICTA-SIGN:} sign language recognition, generation and modelling
  with application in deaf communication.
\newblock {\em International workshop on the Representation and Processing of
  Sign Languages: Corpora and Sign Language Technologies ({LREC)}, Valleta,
  Malte}, pages 80--83.

\bibitem[\protect\citename{Filhol}2009]{filhol_descriptive_2009}
Michael Filhol.
\newblock 2009.
\newblock A descriptive model of signs for sign language processing.
\newblock {\em Sign Language \& Linguistics}, 12(1):93--100.

\bibitem[\protect\citename{Filhol}2012]{filhol_combining_2012}
Michael Filhol.
\newblock 2012.
\newblock Combining two synchronisation methods in a linguistic model to
  describe sign language.
\newblock In Eleni Efthimiou, Georgios Kouroupetroglou, and Stavroula-Evita
  Fotinea, editors, {\em Gesture and Sign Language in Human-Computer
  Interaction and Embodied Communication}, number 7206 in Lecture Notes in
  Computer Science, pages 194--203. Springer Berlin Heidelberg, January.

\bibitem[\protect\citename{Gonzalez and Collet}2011]{workshop_signs_2011}
Matilde Gonzalez and Christophe Collet.
\newblock 2011.
\newblock Signs segmentation using dynamics and hand configuration for
  semi-automatic annotation of sign language corpora.
\newblock {\em Gesture in Embodied Communication and Humain-Computer
  Interaction}, pages 100--103, May.

\bibitem[\protect\citename{Liddell}2003]{liddell_grammar_2003}
Scott~K. Liddell.
\newblock 2003.
\newblock {\em Grammar, Gesture, and Meaning in American Sign Language}.
\newblock Cambridge University Press, March.

\bibitem[\protect\citename{Mahanti \bgroup et al.\egroup
  }2003]{mahanti_framework_2003}
Ambuj Mahanti, Supriyo Ghose, and Samir~K. Sadhukhan.
\newblock 2003.
\newblock A framework for searching {AND/OR} graphs with cycles.
\newblock {\em {arXiv} preprint cs/0305001}.

\bibitem[\protect\citename{Neidle \bgroup et al.\egroup
  }2009]{neidle_method_2009}
Carol Neidle, Joan Nash, Nicholas Michael, and Dimitris Metaxas.
\newblock 2009.
\newblock A method for recognition of grammatically significant head movements
  and facial expressions, developed through use of a linguistically annotated
  video corpus.
\newblock In {\em Proceedings of the Language and Logic Workshop, Formal
  Approaches to Sign Languages, European Summer School in Logic, Language, and
  Information ({ESSLLI} 2009)}, Bordeaux, France.

\bibitem[\protect\citename{Paulraj \bgroup et al.\egroup
  }2008]{paulraj_extraction_2008}
M.~P. Paulraj, Sazali Yaacob, Hazry Desa, C.~R. Hema, and Wan~Ab Majid.
\newblock 2008.
\newblock Extraction of head and hand gesture features for recognition of sign
  language.
\newblock In {\em Proc. International Conference on Electronic Design {ICED}
  2008}, pages 1--6.

\bibitem[\protect\citename{Vermeerbergen \bgroup et al.\egroup
  }2007]{vermeerbergen_simultaneity_2007}
Myriam Vermeerbergen, Lorraine Leeson, and Onno~Alex Crasborn.
\newblock 2007.
\newblock {\em Simultaneity in Signed Languages: Form and Function}.
\newblock John Benjamins Publishing, January.

\bibitem[\protect\citename{Yang and Lee}2011]{yang_combination_2011}
Hee-Deok Yang and Seong-Whan Lee.
\newblock 2011.
\newblock Combination of manual and non-manual features for sign language
  recognition based on conditional random field and active appearance model.
\newblock In {\em 2011 International Conference on Machine Learning and
  Cybernetics ({ICMLC)}}, volume~4, pages 1726--1731.

\end{thebibliography}

\end{document}